# TEXTURE IMAGE ANALYSIS AND TEXTURE CLASSIFICATION METHODS - A REVIEW

Laleh Armi[1], Shervan Fekri-Ershad[1,*]

[1] *Faculty of Computer Engineering, Najafabad Branch, Islamic Azad University, Najafabad, Iran*

[*]*Corresponding Author: fekriershad@pco.iaun.ac.ir*

## Abstract

Tactile texture refers to the tangible feel of a surface and visual texture refers to see the shape or contents of the image. In the image processing, the texture can be defined as a function of spatial variation of the brightness intensity of the pixels. Texture is the main term used to define objects or concepts of a given image. Texture analysis plays an important role in computer vision cases such as object recognition, surface defect detection, pattern recognition, medical image analysis, etc. Since now many approaches have been proposed to describe texture images accurately. Texture analysis methods usually are classified into four categories: statistical methods, structural, model-based and transform- based methods. This paper discusses the various methods used for texture or analysis in details. New researches shows the power of combinational methods for texture analysis, which can't be in specific category. This paper provides a review on well known combinational methods in a specific section with details. This paper counts advantages and disadvantages of well-known texture image descriptors in the result part. Main focus in all of the survived methods is on discrimination performance, computational complexity and resistance to challenges such as noise, rotation, etc. A brief review is also made on the common classifiers used for texture image classification. Also, a survey on texture image benchmark datasets is included.

## I.  INTRODUCTION

The texture is recognizable in both tactile and optical ways. Tactile texture refers to the tangible feel of a surface and visual texture refers to see the shape or contents of the image [1]. Diagnosis of texture in a human vision system is easily feasible but in the machine vision domain and image processing have their own complexity. In the image processing, the texture can be defined as a function of spatial variation of the brightness intensity of the pixels[2]. The texture represents the variations of each level, which measures characteristics such as smoothness, smoothness, coarseness and regularity of each surface in different order directions. Textural images in the image processing and machine vision refer to the images in which a specific pattern of distribution and dispersion of the intensity of the pixel illumination is repeated sequentially throughout the image[2].

In the following Fig, a type of natural image with repetitive texture is represented. Fig 1.a  illustrates the image of a wall of wood with a fully repetitive texture. This pattern is repeated throughout the image. The bamboo pattern is also shown in Fig 1.b.

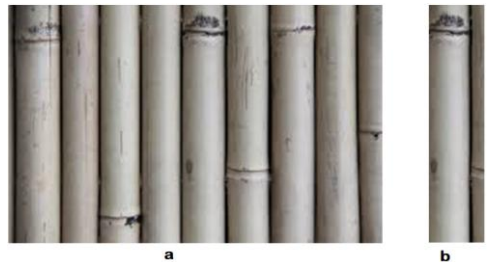

*Fig 1: An example of a textural image: a) texture Image; b) Repetitive pattern of texture*

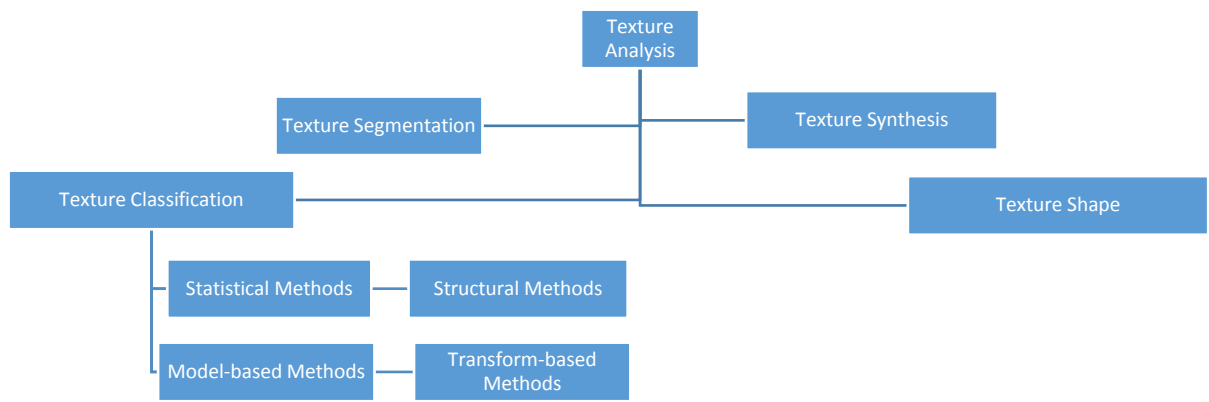

*Fig 2:Texture analysis areas in a general view*

As it is indicted in the Fig 2, texture classification, texture segmentation, texture synthesis, and texture shape are among the main issues that texture analysis deals with [1-3].

In the "Texture Shape Extraction", the objective is to extract 3D images which are covered in a picture with a specific texture. This field studies the structure and shape of the elements in the image by analyzing their textual properties and the spatial relationship each with each other.

The purpose of "Texture Synthesis" is to produce images that have the same texture as the input texture. Applications of this field are creation of graphic images and computer games. Eliminate of a part of the image and stow it with the background texture, creation of a scene with lighting and a different viewing angle, creation of artistic effects on images like embossed textures are other applications of this field.

The purpose of the "Texture Segmentation" is to divide an image into distinct areas, each of which is different in terms of texture. The boundaries of different textures are determined in the texture segmentation. In other words, in texture segmentation, the features of the boundaries and areas are compared and if their texture characteristics are sufficiently different, the boundary range has been found.

Texture classification is one of the important areas in the context of texture analysis whose main purpose is to provide descriptors for categorizing textural images. Texture classification means assigning an unknown sample image to one of the predefined texture classes. The key issues in the analysis of texture are texture classification which will be dealt with in this article.

A.    *Challenges in Texture Images:*

In real condition, there are two major challenges in the analysis and classification of images which has many destructive effects. These two important phenomena are rotation and noise image. If the methods used to classify against these common phenomena are not sustainable, in practice, the accuracy of the results can be severely reduced; therefore, in actual circumstances, the methods used to analyze and categorize the images should be as robust and stable as possible to these two phenomena and neutralize their devastating effects. In addition to the above issues, the images may differ from one another in terms of scale, viewpoint or intensity of light. This is one of the leading challenges in the texture classification system. For this reason, various methods have been proposed, each of which tries to cover these aspects.

B.    *Feature Extraction Method for Categorizing Textures:*

As mentioned above, the texture classification means assignment of a sample image to a previously defined texture group. This classification usually involves a two-step process.

   A)    The first stage, the feature extraction phase:
In this section, textural properties are extracted. The goal is to create a model for each one of the textures that exist in the training platform.

   B)    The second stage, the classification phase:
In this phase, the test sample image texture is first analyzed using the same technique used in the previous step and then, using a classification algorithm, the extraction features of the test image are compared with the train imagery and its class is determined. The general flowchart of methods for the texture images classification is indicated in Fig 3, based on the two preceding phases

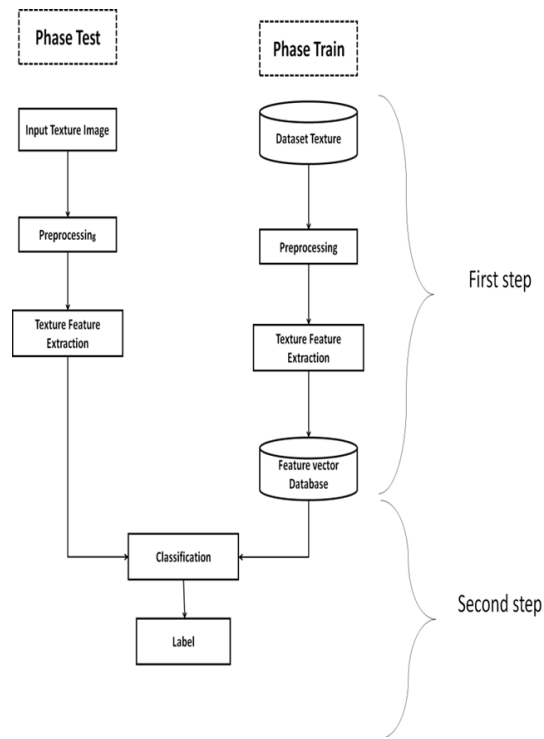

Fig 3: The popular flowchart of the texture images classification process

*I.B.1    Feature Extraction Phase:*

The first stage in extracting texture features is to create a model for each one of the textures found in educational imagery. Extractive features at this stage can be numerical, discrete histograms, empirical distributions, texture features such as contrast, spatial structure, direction, etc. These features are used for teaching classification. So far, many ways to categorize texture have been proposed which the efficiency of these methods depends to a great extent on the type of features extracted. Among the most common ones, they can be divided into four main groups of "statistical methods", "structural methods", "model-based methods", "transform methods" Each of these methods extracts the various features of the texture [3, 4]
It is worth noting that today it is difficult to put some methods in a particular group due to the complexity of the methods and the use of the combined properties because most of them fall into several groups.
Types of widely used and popular methods of features extraction texture will be described in detail in the next section.

*I.B.2    Classification Phase:*

In the second stage, the texture classification is based on the use of machine learning algorithms with monitoring or classification algorithms; so that the appropriate class for each image is selected from the comparison of the vector of the extracted features in the educational phase with the vector of the selection test phase characteristics and its class is determined. This step is repeated for each image that is in the test phase. At the end, the estimated classes for testing with their actual class are adapted and the recognition rate of each algorithm is calculated which indicates the efficiency of the implemented method which the recognition rate of each algorithm is used to compare the efficiency of its algorithm with other available methods.

$$Classification\ accuracy = \frac{Number\ of\ correct\ Matches}{Total\ Number\ of\ test\ images} \times 100\% \quad (1)$$

*C.    Texture Analysis Application:*

The image texture gives us a lot of useful information about the content of the image, the objects inside it, the background context, background, and so on. Texture analysis in most areas of image processing, especially in the process of learning and extracting the feature is being discussed when we want to compare the images such as:

- Face Detection [5-8]
- Tracking Objects in Videos [9]
- Product Quality Diagnostics[10]
- Medical image analysis[11, 12]
- Remote Sensing[13]
- Vegetation[14]

*D.    Paper Organization:*

This article has been compiled in 7 sections. In the second section, we review the four categories of texture classification and some of the many methods and features used to describe the texture. After reviewing the fundamental methods in this area, the combined methods of texture analysis are discussed in the third section. The fourth section of the well-known texture data set is introduced. In the following, we describe several efficient classifier algorithms including K-nearest neighbor, artificial neural networks, support vector machines, and so on. In Section six, a general overview of the data used in the textures and the summary of the methods are gathered with their advantages and disadvantages. The discussion and conclusion will be represented in the seventh section.

## II.    Texture Classification Methods:

In this section, the definition of the concept of each group (statistical, structural, model-based, transform) is discussed and then, there are several ways to deal with it. Table 1 indicates a list of some of the methods presented in this field.

Table 1: Categorization of texture classification methods

| Categories | Sub-categories | Method |
|---|---|---|
| Statistical | Histogram Properties | • Binary Gabor pattern[15] <br> • GLCM and Gabor filters[10] <br> • Gabor and LBP[16] <br> • wavelet transform and GLCM[17] <br> • local binary patterns and significant point's selection [18] <br> • Energy variation[4] <br> • Combination of primitive pattern units and statistical features[19] <br> • Hybrid color local binary patterns  [70] |
|  | Co-occurrence Matrix |  |
|  | Local Binary Descriptors |  |
|  | Registration- based |  |
|  | Laws Texture Energy |  |
| Structural | Primitive Measurement | • Energy variation[4] <br> • Edge-based texture granularity detection[20] <br> • Morphological filter <br> • Skeleton primitive and wavelets |
|  | Edge Features |  |
|  | Skeleton  Representation |  |
|  | Morphological Operations |  |
|  | SIFT |  |
| Model-based | Autoregressive (AR) model | • Multifractal Analysis in Multi-orientation Wavelet Pyramid[21] <br> • Markov Random Field Texture Models [22] <br> • simultaneous autoregressive models[23] |
|  | Fractal models |  |
|  | Random Field Model |  |
|  | Texem Model |  |
| Transform-based | Spectral | • Binary Gabor Pattern[15] <br> • wavelet channel combining and LL channel filter bank[24] <br> • GLCM and Gabor Filters[10] <br> • Gabor and LBP[16] <br> • wavelet transform and GLCM[17] <br> • SVD and DWT domains[25] <br> • Skeleton primitive and wavelets |
|  | Gabor |  |
|  | wavelet |  |
|  | Curvelet Transform |  |

### A.    Statistical Methods:

The methods of statistical processing of texture that make up many of the methods presented in the machine vision field, spot localization of pixel values. These batches of methods for analyzing the texture of images perform a series of statistical calculations on the lightness intensity distribution functions of pixels. In general, the methods used to derive feature vector from statistical computations fall into this group. The first, second and higher level statistical characteristics are among these methods. The difference between these type of three features is that the first- level single-pixel specification is calculated without taking into account the interaction between pixels of the image. While in a second-level and higher- level statistical characteristic, the specification is calculated taking into account the dependence of two or more pixels. The Co-Occurrence Matrix that is known as the second-level histogram is one of the methods to be included in this group.

### II.A.1    Histogram Features and Specifications:

First-order statistical indicators are calculated directly from the gray levels of the pixels of the original image, regardless of the spatial relationship between them. Typically, the first- level statistical indexes are derived from the calculation of the statistical moments of the histogram of the image. The image histogram is a two-dimensional representation of how the gray levels are dispersed in the image. Simply, histogram is a graphical representation. It shows us the optical content of the image. The meaning of optical content is the amount of light and darkness of the image.

One of the easiest methods to describe a texture is the use of statistical torques related to the histogram of the intensity of an image or region, and many features can be extracted from it which is used to calculate the nth torque around the mean as follows:

$$\mu_n(z) = \sum_{i=0}^{L-1}(z_i - m)^n p(z_i) \qquad (2)$$

$$m = \sum_{i=0}^{L-1} z_i p(z_i) \qquad (3)$$

$z_i$ is the random variables related to the intensity rang of the image, $p(z_i)$ represents the number of pixels in the image that have a gray level equal to $z_i$.

The variance is a measure of contrast severity, which can be used to create descriptors of relative smoothness.

$$\sigma^2 = \sum_{i=0}^{L-1}(z_i - m)^2 p(z_i) \qquad (4)$$

Skewness: A criterion for the histogram symmetry degree

$$skewness = \sum_{i=0}^{L-1}(z_i - m)^3 p(z_i) \qquad (5)$$

Entropy: A criterion for variability and is zero for a fixed image.

$$\text{Entropy} = -\sum_{i=0}^{L-1} p(z_i) \log_2 p(z_i) \qquad (6)$$

Fig 4 is a histogram of 2 pond images with a homogenized surface; ruggedly illustrated.

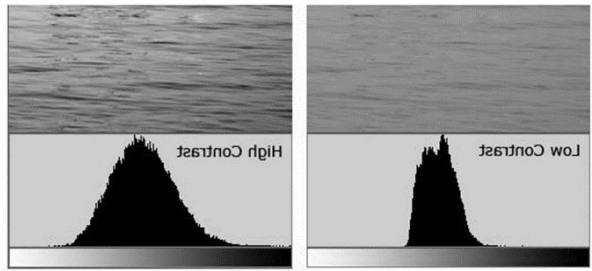

Fig 4: An example of the histogram computing  a) The image of the river with a homogenized surface b) A image histogram c) A river image with a rugged surface d) A image histogram

*II.A.2    Co-Occurrence Matrix:*

One of the oldest operators for extracting texture features is the co-occurrence matrix introduced by Haralick [26]. The co-occurrence matrix of an image is created based on the correlations between image pixels. For a k-bit image with $L = 2k$ brightness levels, an $L \times L$ matrix is created whose elements are the number of occurrences of a pair of pixels with brightness of a.b separated by d pixels in a certain direction. After calculating the matrix, the textual characteristics of the second statistic (Haralick) is calculated.

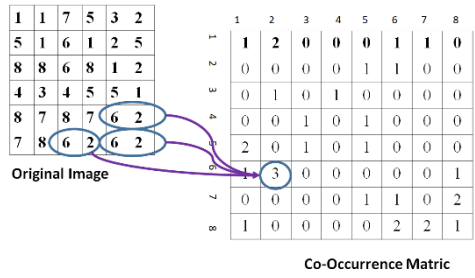

.

Fig 5: An Example of How to Extract a Co-Occurrence Matrix with 8 Brightness Levels

For example, in Fig 5, a 3-bit image with 8 levels of intensity is shown which its co-occurrence matrix has 8 rows and 8 columns. The elements of this matrix are the number of pixel occurrences with gray levels $i.j$ which are represented by a displacement of 1 pixel in the direction of zero degrees.

Typically, the co-occurrence matrix is defined for the four main directions (0, 45, 90, and 135). In Fig 6, four possible angles between two pixels with angles (0, 45, 90 and 135) degrees are represented with a displacement of 3 between two pixels.

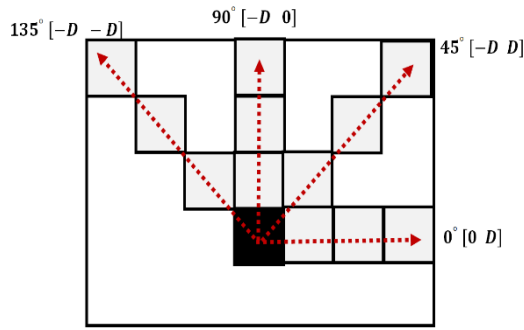

Fig 6: Four different directions with displacement 3 between two pixels

After the formation of the co-occurrence matrix, the Haralick statistical properties can be calculated from the output of the co-occurrence matrix. For example, 6 statistical properties (Contrast, correlation, energy, homogeneity, Entropy, and maximum probability) can be deduced from the co-occurrence matrix.

$$\text{Contrast} = \sum_{i,j} |i - i|^2 p_{i,j} \qquad (7)$$

$$\text{Correlation} = \sum_{i,j} \frac{(i - \mu_i)(j - \mu_j) p_{i,j}}{\sigma_i \sigma_j} \qquad (8)$$

$$\text{energy} = \sum_{i,j} p_{i,j}{}^2 \qquad (9)$$

$$\text{homogeneit} = \sum_{i,j} \frac{p_{i,j}}{1 + |i - j|} \qquad (10)$$

$$\text{Entropy} = -\sum_{i,j} p_{i,j} \, log_2 \, p_{i,j} \qquad (11)$$

$$\text{maximum probability} = \max(p_{i,j}) \qquad (12)$$

In the above equations, the $p_{i,j}$ is the normalized co-occurrence matrix and $\mu_i$ quantity is an average that is calculated along rows of matrixes $G$ and $\mu_j$ is the average that is calculated along the columns. Similarly, $\sigma_i$ and $\sigma_j$ are standard deviations which are calculated along rows and columns.

### II.A.3 Local Binary Patterns:

Local binary pattern is one of the textural image descriptors that can define the local spatial structure and the local contrast of the image or part of that. LBP has become a widely used texture descriptor due to its simple implementation and extraction of proper features with high classification accuracy, and this is why many researchers have considered it. An important feature of this method, i.e., its sustainability in the uniform changes of the gray scale and computational efficiency, has made it one of the most suitable image analysis methods. Ojala et al first introduced the LBP operator.[27] This descriptor is independent of the rotation. In this method, a neighborhood is considered first for each point of the image. Then the intensity of the central pixel is compared with the intensity of the neighboring pixels. If the intensity of the neighboring pixel illumination is larger than the central pixel, then the value for that neighbor in extracted binary pattern considered as one, and otherwise it is zero. Finally, with a binary weighted sum of the values in the binary extraction pattern we obtain values at the base of ten, which is called the LBP value. The calculation of the final LBP is shown in the equation below.

$$LBP_{P,R} = \sum_{p=0}^{P-1} S(g_p - g_c) 2^p \qquad (13)$$

$$S(x) = \begin{cases} 1 & g_p - g_c > 0 \\ 0 & otherwise \end{cases} \qquad (14)$$

P in Eq.13 represents the number of neighbors for central pixel, $g_c$ represents the intensity of the central pixel illumination and $g_p$ represents the illumination intensity of the P-th neighbor pixel. The $S(x)$ sign function is used to generate binary values, and is calculated according to Eq. 14. This operator is applied to all pixels of the image and eventually the histogram of LBP values is constructed that can be used to extract the features. An example of how to calculate a LBP shown in Fig. 7.

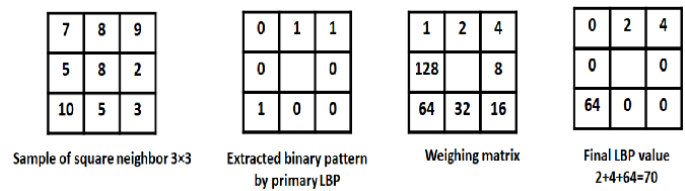

*Fig 7: How to calculate the LBP value in a 3*3 neighborhood*

In order to prevent the operator from being sensitive to rotation, the neighborhood is considered as a circle and the illumination intensity of the points whose coordinates do not exactly match the center of the pixel is determined through interpolation. Fig.8 shows a circular neighborhood with different radius R and neighboring points number P.

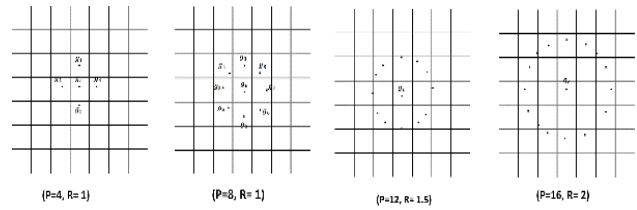

Fig 8: Circular symmetric neighbors for different values of P and R[29]

Given the definition provided for the LBP, the output of this operator is a binary P-bit number so that $2^P$ has different values. When an image is rotated, all neighbors will rotate around the central pixel $g_c$ in one direction, and this will result in the production of different values for $LBP_{P,R}$. In other words, the value of $LBP_{P,R}$ depends on the indexing of neighboring pixels. To eliminate rotating destructive effects, the bit rotation to right operator is used so that a unique value is assigned to each of the patterns [28]. This independent rotation operator is calculated as follows

$$LBP_{P,R}{}^{ri}(x,y) = min\{ROR( LBP_{P,R}(x,y), i) | i \in [0, P-1]\} \quad (15)$$

In Eq. 15, the $ROR$ function denotes the rotation to the right on the P-bits binary number, and the $ri$ symbol represents the operator's insensitivity to the rotation. This is performed $i$ times and the minimum number obtained for each $i$ between $0$ to $P-1$ is the final value of the LBP.

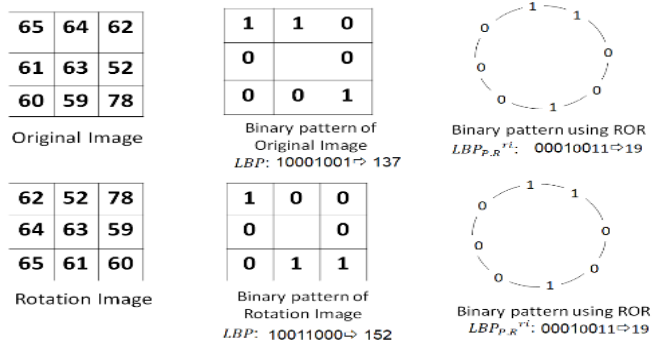

Fig 9: A numeric example of LBP process using the bit rotation to the right ($LBP_{P,R}{}^{ri}$)

As indicated in the $LBP_{P,R}{}^{ri}$ method, the resulting binary number is rotated and the minimum value is assigned as the local pattern for the neighboring center of interest. This method, in its original form, has a high computational complexity and, with increasing number of neighboring points, this computational complexity increases. Many modifications are proposed to improve LBP. The first modification for LBP, called the uniform MLBP, was introduced by Ojala [29].

In MLBP, the uniformity shows the number of mutations occurring between 0 and 1 (and vice versa) in the binary pattern extracted from neighbors, and is represented by the symbol '$U$'. For example, the binary numbers 00000000 and 11111111 show no mutation and the binary number 11010110 shows 6 mutations. In this way, the modified LBP operator is rotation-insensitive In Eq. 16, the number of mutants between the bits is defined. In MLBP, patterns with homogeneity less than or equal to $U_T$ are defined as uniform patterns, and patterns with homogeneity greater than $U_T$ are defined as heterogeneous patterns. Given Eq. 17, the labels 0 to $P$ are assigned to homogeneous neighbors and the $P+1$ is assigned to heterogeneous neighbor. After applying this operator to textural images and assigning different labels, the probability of a particular label can be approximated by the ration of the number of labeled points to the total number of points in the entire image. So eventually, a probability eigenvector of $P + 2$ for each input image is obtained. In Eq. 17, according to references[29], $U_T$ is usually considered to be $P/4$, and only a small amount of patterns in the texture is labeled as $P + 1$.

$$U(LBP) = |s(g_{P-1} - g_c) - s(g_0 - g_c)| + \sum_{p=0}^{P-1} |s(g_p - g_c) - s(g_{p-1} - g_c)| \quad (16)$$

$$MLBP_{P,R}{}^{riu2} = \begin{cases} \sum_{p=0}^{P-1} s(g_p - g_c) & if \ U(LBP) \leq U_T \\ P + 1 & otherwise \end{cases} \quad (17)$$

### B. Structural Methods:

In this set of methods, the texture is introduced based on the initial units and their spatial layout. Initial units can be simply as a pixel, a region, or a line like shape. Their spatial layout rules are arranged together by calculating geometric relationships or examining their statistical properties. In other words, structural methods consider texture as a combination of initial patterns, once the primary texture is detected, and then the statistical properties of the primary texture are calculated, and used as a feature. These methods are suitable for textures with a regular structure but for images with irregular texture, is not optimal method.

### II.B.1 Edge Features:

The edge of the border is between an object and its background, in other words, it is the edges of the two gray levels or the values of the two-pixel brightness adjacent which occurs in a specific place of the image. The purpose of the edge detection is to identify the points of an image in which the intensity of light shifts sharply. The edges may be subject to vision. That is, they can change by changing the point of view, and, typically, the geometry of the scene, the objects that intercept each other and so on, or may be affiliated with the perspective - Which usually represents the features of the objects seen, such as markings and surface shape. This action can be done by various operators like Sobel[30], Pewit[31], Robert and etc.

### II.B.2 Scale Invariant Feature Transform (SIFT):

SIFT operator today is one of the best and most powerful tools for extracting and describing features. This method was introduced by David Loo in 1999[32]. The advantage of this operator is that it has local features therefore, it has good accuracy in occlusion it is differentiable. In the following Fig, the process of extracting features is indicated by the SIFT operator. In the first step, the key points of the image are identified. The key points of the image are the points of the image that Difference of Gaussian (DoG) is maximized or minimized in those points[33].

The process of finding these points begins by constructing a pyramid of images and convolution of the original image with the Gaussian filters $G(x, y, \sigma)$ therefore the scale space is displayed as follows[33].

- Scale-space extrema detection
- key point localization
- Orientation assignment
- Key point descriptor

The maturity level is controlled by the standard deviation parameter σ in the Gaussian function. The DOG scale space is also achieved by subtracting adjacent scale levels. Fig 10 below indicates the DoG construction steps.

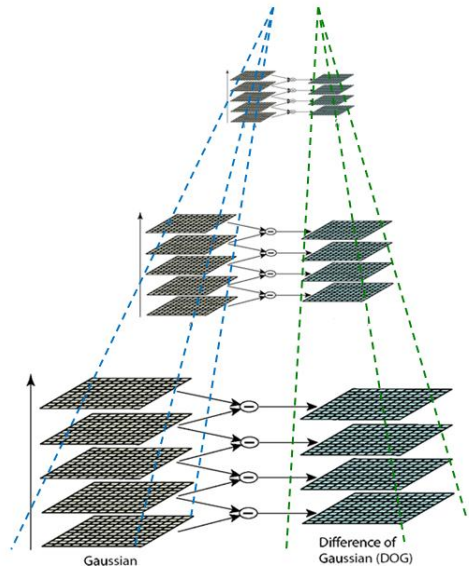

Fig 10: Construction of DoG image

The next step is to find the maximum or minimum points in each octave. This is done by comparing each pixel with a neighbor 26th in the $3 \times 3$ region of all adjacent DoG levels in the same octave. If the desired point is larger or smaller than all its neighbors, it is chosen as the target point[33].

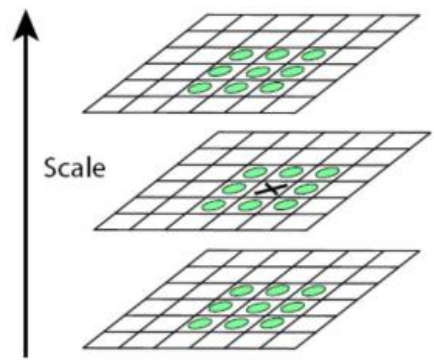

Fig 11: Maxima and Minima Identification of DoG Images[33]

The next step is to remove points with low and unstable contrast and points on the edge. At the key point descriptor stage, the main property vector will be extracted. Initially, the gradient domain and the path around the key point are sampled using the $4 \times 4$ array with 8 directions for the histogram. Therefore, the vector length is 128 for each point. David loo in[33] called extraction property vectors, SIFT points.

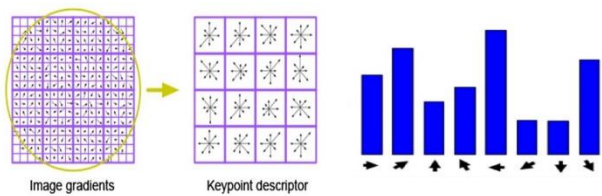

Fig 12: Creation of Keypoint Descriptor[33]

### C.      Model-Based Methods:

Model-based methods on the basis of model design can describe texture images. The model-based method is used for texture modeling, and the most popular ones are Autoregressive (AR) method, Markov Square theory, Gibbs RMF, Hidden Markov Model(HMM), and fractal model. In this way, a model of the image is created and then this model is used to describe the image synthesis. Model parameters extract the basic qualitative properties of the texture. A fractal geometric model is a model used to analyze many natural and physical phenomena.

### II.C.1     Fractal:

In 1970 Mendelbert introduced a new field of mathematics called fractal to the world[34]. He introduced a new class of collections called fractal. This collection contains many complicated objects which were produced by repeating simple rules. The fractals can be used to model the coarseness and harshness and self-similarity in a texture image. This feature is in the category of model-based methods. If a collection $A$ of repetition of $N$ is a separate copy of itself, the collection $A$ is called self-similar. Each of these copies is shifted to $r$ from the original image. The fractal dimension in Fig 13. D is obtained from the interface between $N$ and $r$.

$$D = log \frac{N}{log\left(\frac{1}{r}\right)} \tag{18}$$

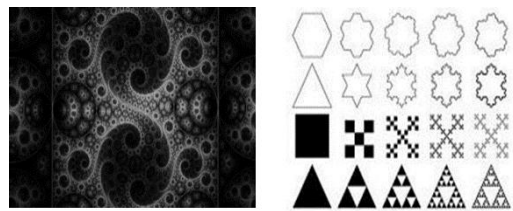

Fig 13: An Example of a Fractal Model

### D.      Transform-based Methods:

In these methods, the image is transformed into another space (using a transformation function), so that the texture is more easily distinguishable in the new space. The most commonly used extraction methods in this method are Wavelet transforms[35], Curvelet[36], Ridgelet [37] and Gabor [38]

### II.D.1     Spectral Measures of Texture:

The Fourier spectrum is suitable for describing the alternate patterns or nearly alternating two-dimensional patterns in an image. These general texture patterns can be well-identified as concentrations above the energy in the spectrum. Three features of the Fourier spectrum that are useful in describing texture are as follows [39]

• Peak arrives in the main direction pattern of the texture pattern

• Peak on the main alternate frequency plate of the pattern location

• Elimination of any proportional components by filtering non-alternate image components

The spectrum around the source is symmetric, so only half of the frequency plate is considered. Therefore, for the analysis of each alternate pattern, only one peak is related to the spectrum. It is usually performed by displaying the spectrum of polar coordinates as a function $s(r, \theta)$ where s is the function of the spectrum and r is the variables of this coordinate system. For each, $\theta$ and $s(r, \theta)$ can be considered as one dimension of $s_\theta(r)$. Similarly, for each frequency r, $s_r(\theta)$ is a one-dimensional function. The analysis of $s_\theta(r)$ for a constant value of $\theta$ shows the behavior of the spectrum (the presence of peaks in the spectrum) along the radial direction of the origin while a more general description is obtained by integration of this function, it is easier to distinguish and interpret spectral properties by expressing the spectrum in polar coordinates. S is the spectral function; r and $\theta$ are variables in polar coordinates. R0 is the radius of the circle corresponding to the origin; S (r) is constant toward the rotation[39].

$$R_0 = \text{Round} \quad \left(\min \frac{m,n}{2}\right) - 1 \qquad (19)$$

$$S(\theta) = \sum_{i=0}^{R_0} S_r(\theta) \quad 1 \leq r \leq 180 \qquad (20)$$

$$S(r) = \sum_{i=0}^{R_0} S_\theta(r) \quad 1 \leq r \leq R_0 \qquad (21)$$

*II.D.2    Gabor:*

One of the considerable methods in the texture analysis is transform method. The Gabor transform is similar to the wavelet transform, in which functions have the Gaussian nature base and as a result, this transformation is optimal in the frequency domain arrangement [40] Gabor wavelet is an optimal transform to minimize the two dimensional uncertainty associated with the location and frequency domains. This wavelet can be used as directional and comparable scale detectors for revealing lines and edges in images [41]. Also, the statistical properties of this transformation can be used to determine the structure and visual content othe f images. The features of Gabor's transformation are used in several applications of image analysis, including categorization and texture segmentation, image recognition, alphabet recognition, image recording, and routing and movement. Two-dimensional Gabor filters are defined in the spatial and frequency domains.

A two-dimensional Gabor consists of a Gaussian modulus function with a mixed sinusoidal function. This function can be expressed in terms of $\sigma_y$ and $\sigma_x$ and the standard deviation $\psi$ of the sinusoidal function of the Gaussian function is as follows:

$$G = exp\left(-\frac{x'^2 + \gamma^2 y^2}{2\sigma^2}\right) exp\left(i\left(2\pi \frac{x'}{\lambda} + \psi\right)\right) \qquad (22)$$

$$G = exp\left(-\frac{x'^2 + \gamma^2 y^2}{2\sigma^2}\right)\left(cos\left(2\pi \frac{x'}{\lambda} + \psi\right) + sin\left(2\pi \frac{x'}{\lambda} + \psi\right)i\right) \qquad (23)$$

$$G = \left(exp\left(-\frac{x'^2 + \gamma^2 y^2}{2\sigma^2}\right)cos\left(2\pi \frac{x'}{\lambda} + \psi\right) + exp\left(-\frac{x'^2 + \gamma^2 y^2}{2\sigma^2}\right)sin\left(2\pi \frac{x'}{\lambda} + \psi\right)i\right) \qquad (24)$$

$$x' = x\,cos\theta + y\,sin\theta \quad AND \quad y' = -x\,sin\theta + y\,cos\theta \qquad (25)$$

$$G = \frac{e^{\frac{\left(\frac{x-x_0}{\sigma_x}\right)^2 \left(\frac{y-y_0}{\sigma_y}\right)^2}{2}} e^{j[u_0(x-x_0) + v_0(y-y_0)]}}{2\pi\sigma_x\sigma_y} \qquad (26)$$

$\sigma_x$ and $\sigma_y$, respectively, determine the spatial coordinates of the Gaussian function in line with $x.y$. $(x_0.y_0)$ Is function center and $(u_0.v_0)$ is the frequency center of the function. Different applications of the computer vision, such as texture analysis and edge detection, Gabor filter has been used extensively. The Gabor filter is a linear and local filter. The convolutional nucleus of the Gabor filter is the product of a complex exponential and Gaussian function and Gabor filters, if tailored and accurately adjusted, have a very good function in distinguishing the features of the texture and the edges of it. The other feature of Gabor's filters is the high degree of separation above them; this means that their response is completely local and adjustable in the area of the place as well as in the frequency domain. In order that the extracted property of the Gabor filter to be constant over time, and to change the scale and the intensity changes, the Fourier transform is used in addition to using the Gabor filter bank. It is worth noting that the Fourier transform is non-sensitive. Using a Gabor filter is one of the most common filter-based methods for texture extraction. This filter operates in both spatial and frequency domains. In the spatial domain, the core of the filter is obtained from the product of a Gaussian function with a directed sinusoidal function. As a result, the filter produces outstanding responses at points of the image that locally have a certain orientation and frequency. When using the Gabor filter, it can be used to get directions that can be used to match the extracted property of the image to the rotational changes.

# III.  *Combinational state-of-the-art Texture Analysis Algorithms*

### A.  *The Gray Scale Level Co-Occurrence Matrix Method:*

Fernando Roberti and et al. [42] presented a new strategic approach to expand the gray scale level co- occurrence matrix with two different approaches. The first approach is to pyramid the image. That is, the image is considered in five layers and the co-occurrence matrix is obtained on these five images in four directions (0, 45, 90 and 135). In order to extract the features of the image from the five co-occurrence matrix combinations in each direction, 12 characteristics (contrast, correlation, total power, instantaneous inverse difference, total average, total variance, total irregularity, variance difference, difference of irregularity, maximum correlation coefficient) has been extracted. In the end, the combination of the obtained vectors is in each direction, in other words, the resulting response is 48 for this approach. In the second approach, instead of pyramiding the image, we will blur the image into five layers with a Gaussian filter. And then, as in the first turn, 48 statistical attributes are obtained from the combination of the co-occurrence matrix in 4 directions. The two approaches are indicated in the following Fig which a) the first approach and b) the second approach[42].

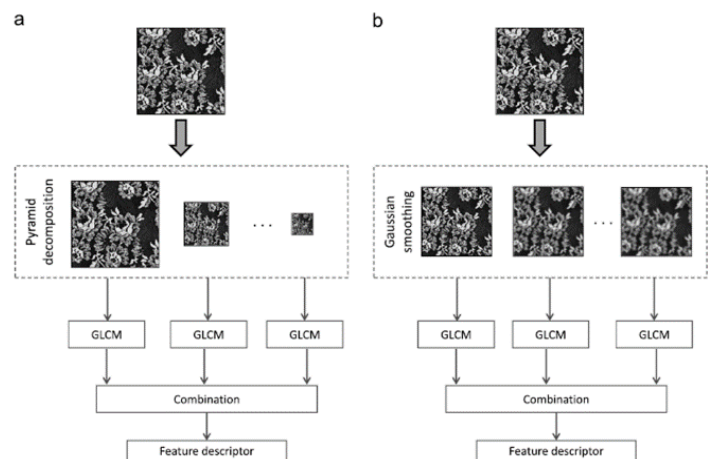

Fig 14: Two presented approaches A) Pyramid image approach b) Image blurring method[42]

### B.  *SEGL Method:*

Fekri Ershad proposed a combinational feature extraction method for textual analysis based on local binary pattern, gray co-Occurrence matrix, edge detection, and statistical features extraction called "SEGL"[43]. Where, term SEGL stands for (Statistical, Edge, GLCM and LBP). Given the flowchart on Fig 15, this method first calculates the initial LBP on the input image. The GLCM is then estimated from the output image from the first sub-box, in 8 directions and sent it to the third box. After using the Sobel filter, the edge of the image entry was briefly identified, 7 statistical features have been extracted.  It consists of entropy, energy, contrast, homogeneity, correlation, average, and variance. In step 2, the co-occurrence matrix is rotated because it is calculated in 8 directions; as a result, the total number of extracted attributes is 56 statistical characteristics[43]. The SEGL method is suggested in Fig 15.

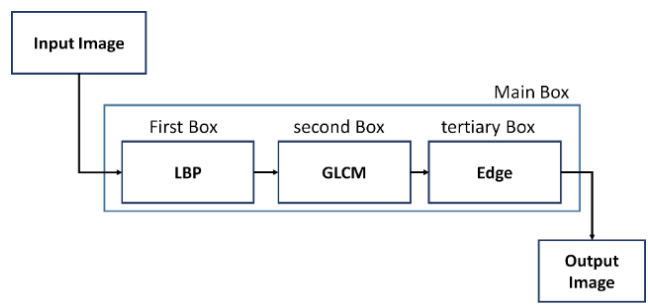

*Fig 15: General Process of Method [43] based on local binary patterns, co-occurrence matrix and edge detection*

*C.    Energy Variation*

In 2012, Fekri Ershad has presented a method for texture classification based on energy variation [4]. In his proposed method, he combines three co-occurrence matrix operators, a local binary pattern, and edge features. In [4], after entrance of the input image, three named operators each one is applied individually to the image and then their energy is extracted. In the next stage from the main input image  a extracted feature vector F [4].

$$F = \langle Energy(LBP), Energy(GLCM), Energy(Edge) \rangle \quad (27)$$

In the following, a characteristic vector provides $F'$, considerable information about the input image can be used to categorize these types of textures. The characteristic vector $F'$ is extracted by finding the difference between the initial energy and the operator output as follows:

$$F' = \langle |O.E - P.E(LBP)|, |O.E - P.E(GLCM)|, |O.E - P.E(Edge)| \rangle \quad (28)$$

Given the above equation, O.E. is original energy and P.E. is the amount of processing energy from the image by operators.

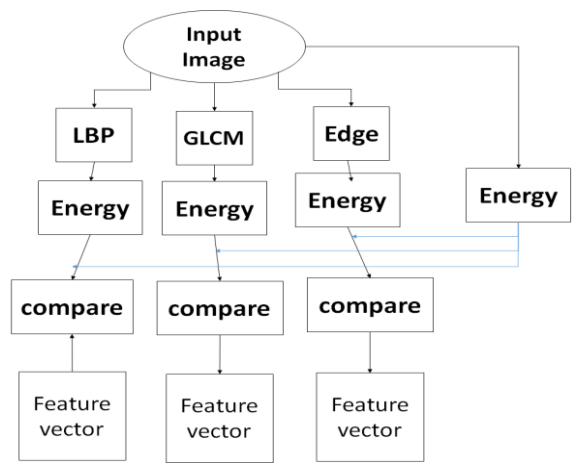

Fig 16: The method presented in[4] based on energy changes

*D.    Binary Gabor Pattern (BGP):*

Zhang et al. have proposed BGP in a as abbreviation binary Gabor Pattern [15] which has been able to improve LBP by using the Gabor filter. The way it works is that it first uses the same eight Gabor filters. These filters are only different in direction. After applying these filters, a piece of eight-dimensional image is obtained. These numbers are binary, but the continuation of the algorithm is similar to the spatially insensitive local binary pattern[15].

$$BGP = \sum_{j=0}^{J} b_j, 2^j \quad (29)$$

$$BGP_{ri} = max\{ROR\,(BGP,j)|j = 0,1, \ldots, J-1\} \quad (30)$$

As the binary string is obtained, the rotational shift is given until it reaches its maximum value. Then each of these bits is multiplied by their respective weight and get together that produces a number. This number is called BGP. For each texture, this value is calculated and then the histogram of the image is obtained. This method uses the closest neighboring algorithm for classifier. The following Fig illustrates an example of applying this method to a different image with different rotation angles. As the results obtained, the two images are the same. This indicates that the operator is resistant to rotation[15].

### E.    Local Spiking Pattern and Application:

Songlin Du et al. in [44] proposed a method that is insensitive to the rotation and intensity of light. In order to extract features they have used a 2-dimensional neural network; which is used in a series of increasing neurons connected to each other and leads to the creation of a binary image. This binary image is encrypted to create a distinctive feature vector. Each pixel in the image of a neuron is increasing in the neural network.

As indicated in the equation below, the output of each neuron in each repetition is a binary value. Depending on the encryption mechanism, it encrypts the output of this network to create a feature vector for texture classification.

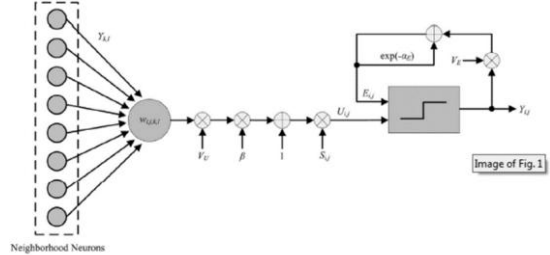

*Fig 17: The spiking cortical neuron model[44]*

### F.    Gray Level Co-Occurrence Matrix Based on Statistical Features:

Soroush et al. [45]have proposed a statistical classification of a texture classification using the method of extraction. There are 2 co-occurrence matrixes in this category. One of these matrixes is derived from the original image and another from the image of the difference. Then it's turn to the statistical characteristics of the contrast, correlation, energy and homogeneity of the two basic matrixes. The feature extraction phase from the proposed classification system based on GLCM is shown in Fig 18.

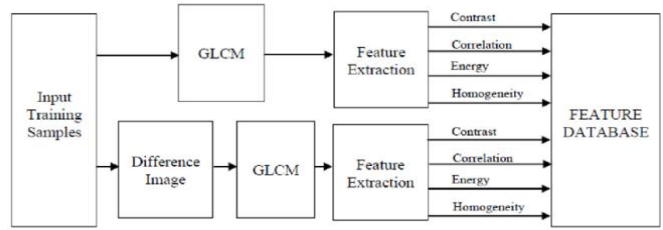

Fig 18:Feature extraction stage of the proposed texture classification method[45]

### G.    Color Texture Classification Approach Based on Combination of Primitive Pattern Units and Statistical Features:

Fekri Ershad proposed another method for color texture classification[19]. In this method, first the, probability occurrence of some primitive pattern units are computed, next, then three histograms are built based on these computation in three different color channels and then the 4 statistical features include Energy, Entropy, Mean and Variance are extracted as features vectors of each channel. Afterward, a total feature vector is derived from a combination these three channel vectors Eq. 31.

$$F_{total} = \langle \begin{matrix} Energy_{HR}, Entropy_{HR}, Mean_{HR}, Variance_{HR}, \\ Energy_{HG}, Entropy_{HG}, Mean_{HG}, Variance_{HG}, \\ Energy_{HB}, Entropy_{HB}, Mean_{HB}, Variance_{HB} \end{matrix} \rangle \qquad (31)$$

Where," HR" is grain components histogram in red channel, "HG" is histogram of grain components in green channel and also "HB" is grain components histogram of color texture in blue channel[19].

### H.     SRITCSD method:

Chang et.al[25] present a new method named the SRITCSD which a rotation-invariant image texture. In this method, first, the Singular Value Decomposition (SVD) is applied to enhance image textures; next, it extracts the texture features in the Discrete Wavelet Transform (DWT) domain in only one level of four sub-bands of: LL1, LH1, HL1 and HH1 the total 0f which produce eight feature values of the SVD version of the image. The SRITCSD method applies the SVM as a multi-classifier for image texture features, where the Particle Swarm Optimization (PSO) algorithm is applied to improve the SRITCSD method. The PSO selects a nearly optimal combination of features and a set of parameters applied in the SVM. In the training and test ases the $f_{DWT}^{j}$ feature is by DWT algorithm as Eq. 32.

$$f_{DWT}^{j} = \left\{ \left( \mu_{l,k}^{j,DWT}, \sigma_{l,k}^{j,DWT} \right) \middle| l = 1, k = 1,2,3,4 \right\} \qquad (32)$$

Where:

$$\mu_{l,k}^{j,DWT} = \frac{\sum_{x=0}^{M_{l,k}-1} \sum_{y=0}^{N_{l,k}-1} F_{DWT}^{j,l,k}(x,y)}{M_{l,k} - 1 \times N_{l,k} - 1} \qquad (33)$$

$$\sigma_{l,k}^{j,DWT} = \frac{\sqrt{\sum_{x=0}^{M_{l,k}-1} \sum_{y=0}^{N_{l,k}-1} \left( F_{DWT}^{j,l,k}(x,y) - \mu_{l,k}^{j,DWT} \right)^2}}{M_{l,k} - 1 \times N_{l,k} - 1} \qquad (34)$$

This is followed by extraction the $f_{SVD;DWT}^{j}$ feature in both phases as Eq. 35.

$$f_{SVD,DWT}^{j} = \left\{ \left( \mu_{l,k}^{j,SVD-DWT}, \sigma_{l,k}^{j,SVD-DWT} \right) \middle| l = 1, k = 1,2,3,4 \right\} \qquad (35)$$

Where:

$$\mu_{l,k}^{j,DWT} = \frac{\sum_{x=0}^{M_{l,k}-1} \sum_{y=0}^{N_{l,k}-1} F_{SVD-DWT}^{j,l,k}(x,y)}{M_{l,k} - 1 \times N_{l,k} - 1} \qquad (36)$$

$$\sigma_{l,k}^{j,DWT} = \frac{\sqrt{\sum_{x=0}^{M_{l,k}-1} \sum_{y=0}^{N_{l,k}-1} \left( F_{SVD\_DWT}^{j,l,k}(x,y) - \mu_{l,k}^{j,SVD-DWT} \right)^2}}{M_{l,k} - 1 \times N_{l,k} - 1} \qquad (37)$$

Finally, the extracted texture feature is obtained from image though combination of the two mentioned steps (DWT-FE and SVD-DWT-FE) as Eq. 38.

$$f^{j} = f_{DWT}^{j} \cup f_{SVD,DWT}^{j} \qquad (38)$$

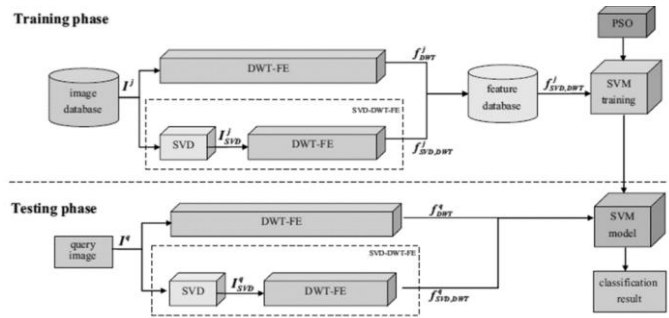

Fig 19: The conceptual design of the SRITCSD method[25]

## IV. Benchmark Texture Image Datasets

Researchers evaluate their proposed algorithm efficiency with standard datasets including the texture image. These images have features that are appropriate for assessing proposed systems and algorithms. Standard dataset provide possibility of comparing algorithms with one other.

*A. Brodatz Texture Album:*

Brodatz [46] is one of the most popular and all-purpose datasets that include natural texture provide by Brodatz through photograph scanned after print. The Brodatz dataset has 112 texture images with resolution of 640×640 pixel and in 8 bit (256 gray values). The Brodatz album has some limitation like single illumination and viewing direction for each one of the texture. Sample images of the Brodatz textures are shown in Fig 20.

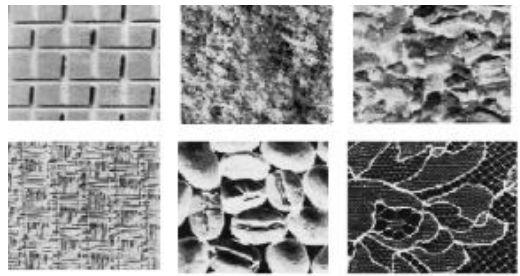

*Fig 20: Examples of texture sample from Brodatz dataset[46]*

*B. Outex Dataset:*

Outex dataset was proposed by Oulue university researcher in 2002. Outex dataset is also most popular and largest dataset. This dataset has both the natural and artificial texture images, which with high applicability like image recovery, texture classification and segmentation problems with open source availability. There 16 sets of texture images among which just Tc_000010 and Tc_000012 are the most different intercity and rotation. These two test suites contain the same 24 classes. OutexTc_000010 has only illuminates "Inca" and OutexTc_000012 have three different illuminates ("Horizon", "Inca" and "TL84"). In this two dataset for each one of different illuminates have nine different rotation angles $(0^\circ, 5^\circ, 10^\circ, 15^\circ, 30^\circ, 45^\circ, 60^\circ, 75^\circ, 90^\circ)$ and 20 textures in each one of rotation angles. Texture images of this dataset are .RAS format

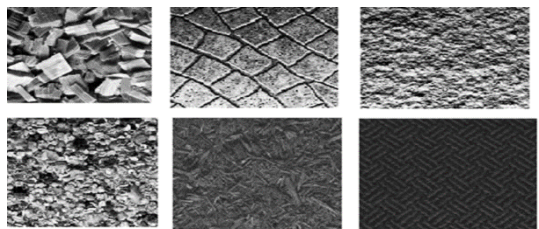

*Fig 21: Examples of texture samples extracted from OuTex dataset[47]*

*C. Vision Texture database (VisTex):*

VisTex dataset was product by MIT University[48]. VisTex database contains color texture images. The goal of VisTex is to provide texture images that are representative of real world conditions. This dataset not only has homogeneous texture image, but it also contains real word scenes of multiple textures. This dataset image is 128×128 pixel size. Each image has special a unique texture pattern. VisTex dataset has 40 classes and whit has 16 samples, that is, a dataset has total of 640 images.

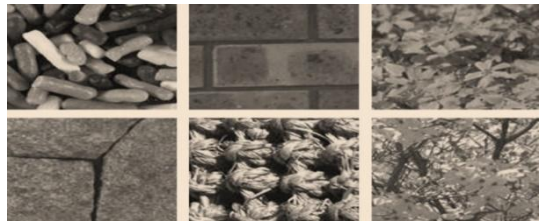

*Fig 22: Examples of texture sample from VisTex dataset [48]*

*D.    Columbia-Utrecht Reflectance and Texture (CUReT) dataset:*

The CUReT dataset is produced in a collaborated research between Columbia University and Utrecht University. This dataset has 61 different classes that each class has 92 images. The size of image is 200×200. Colored images are in .jpg format. These images were provided with different intercity and view point. Therefor this dataset is more suitable than Brodatz for evaluation proposed algorithms [49].

*E.    UMD dataset:*

The UMD (University of Maryland, College Park) is a dataset of high-resolution texture. This dataset with high resolution includes 25 classes and has 40 samples per class. Image size is 960×1280 pixel. This dataset consists of plants, fruits, mosaics etc.

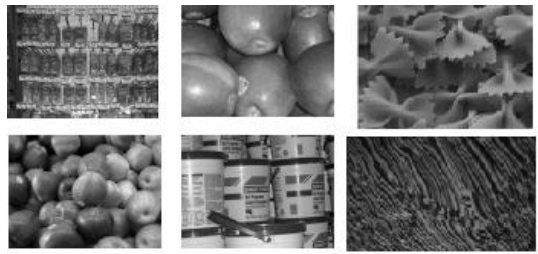

*Fig 23: Examples of texture sample from UMD dataset[50]*

*F.    UIUC dataset:*

This dataset has 25 classes that each class has 40 images and images are 480×480 pixel size. All images in this dataset are in gray surface and have rotate is of different view point and scale [50].

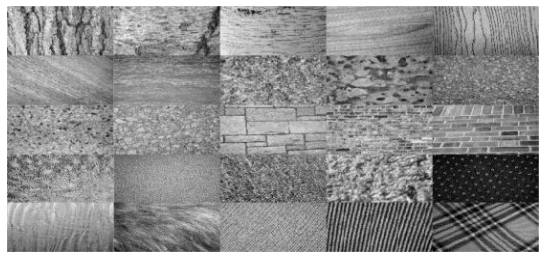

*Fig 24:  Examples of texture sample from UICUC dataset [51]*

*G.    KTH-TIPS dataset:*

This dataset [51] is the exported vermis of CUReT dataset, with Contains 10 texture class images, captured at 9 different scales and 9 different illumination conditions that is, 9×9=81 Images per class. size is 200×200, in .jpg format. Sample images of the KTH-TIPS textures are shown in Fig 25.

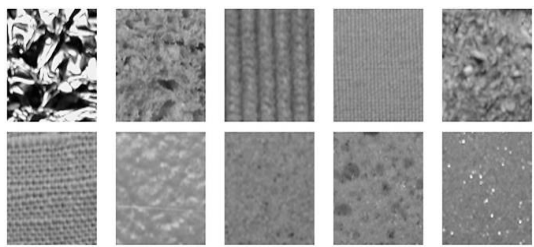

*Fig 25: Examples of texture sample from KTH-TIPS[52]*

*H.     KTH-TIPS2 dataset:*

This dataset is the exported vermis of KTH-TIPS, KTH-TIPS2 has 11 classes with 4 materials per class, and per material has 108 samples that is, each class 432 images a total 4752 of images exist in this dataset. Same image have different colors that result to class in discrimination like the cotton and wool, but some examples of the different classes have identical color for instance cork and biscuit[52]

*I.     Flickr Material Database (FMD)  dataset:*

The FMD dataset was constructed with the specific purpose of capturing a range of real world appearances of common materials include: fabric, foliage, glass, leather, metal, paper, plastic, stone, water, and wood. In this database 100 images (in each category, 50 close-ups and 50 regular views) per category, 10 categories [53].

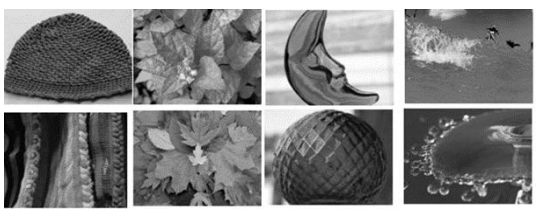

*Fig 26: Examples of texture sample from FMD [53]*

*J.     Other dataset:*

There are a few other datasets that were less popularly used, e.g. Kylberg Texture Dataset[54] , DynTex database[55], TILDA Textile Texture Database[56], Describable Textures Dataset (DTD)[57],  PhoTex (Photometric image set with a wider variation of illumination directions) database[58], Texture library[59] etc.

## V.     Classification Phase in Texture Classification Methods:

In the process of creating a classification system for texture images, after the selection stages of the training set and extraction of features from the images, the model of this categorizer should be created. In order to train this model, machine learning algorithms are used with supervision. For each image, in addition to the feature vector extracted from the image, its class information is also available in the educational set.

*A.     K- Nearest Neighbor:*

The K-nearest neighbor (KNN) algorithm is a sample based and supervised learning method. Just as people try to solve new problems, and use similar solutions that have already been solved; this method in order to classify new data uses the class of previously categorized data. In this method, for each new sample, a separate approximation of the objective function is created; this approximation applies only to the range in the neighborhood of that sample.

The KNN algorithm's performance is assumed initially that all examples of $x_i$  which is having N-dimensional vector, have some vector points in the N-dimensional feature space and k is a positive and determined constant number. In the educational phase, all that is done is to hold the feature vectors and label each educational sample in this N-dimensional space. In the classification phase, the feature vector of the samples whose class is unknown is received as input. Based on the similarity function, k is determined from the educational sample that maps their features closer to the feature vector.

Then the k label of the nearest neighbor of the new sample is being voted and the label that was present in more quantities in this neighborhood would be assigned as new sample. Determining the best value k depends on the problem data. If k is large, it reduces the effect of noise, but in the case of classes with few samples, it isn't being considered and the probability of error increases.

The following is a sample of KNN in Fig 27. Experimental samples are (red circles), educational samples are (green circles and blue circles). In order to specify the experimental sample class to one of the classes (green and red), you must specify k value then assign the educational sample to one of the categories. As you can see in the Fig, if the value of k is 3, then the new sample belongs to the green class, since there are two green circles from the green class and a sample of the blue class and if k is 7, then it will be assigned to the blue class.

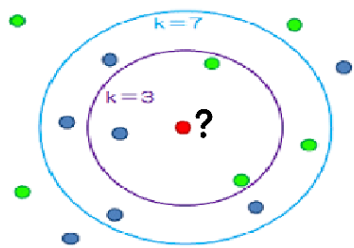

Fig 27: Sample Example of the Nearest Neighbor Algorithm

### B.    Support Vector Machine (SVM):

The support vector machine or SVM was developed by Vapnik [60]. This algorithm is a classification algorithm. In 1995[60], he himself and Mr. Courts generalized for nonlinear states. The support vector machine is a classifier that is part of the Kernel Methods branch in learning the machine [61]. The goal of the SVM algorithm is to identify and distinguish complex patterns or categories of objects in a particular class. In order that SVM split non-linear data, it must use different kernels, it does not work in the 2D space for this rather data is mapped to a more dimensional space so that they can be categorized linearly in this new space. In fact, the main idea behind the support vector machine is to draw clouds in space which performs the differentiation of various data samples in an optimal manner.

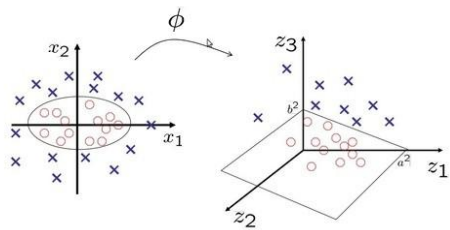

Fig 28: A sample example of SVM  from a two-dimensional to multi-dimensional space

Basically, SVM is a binary separator. In order to assign patterns to several classes, it can be done by combining two-class support vector machines. There are usually two strategies for this goal.

- "1-vs-1" for classification a pair:
In this strategy, we would divide the data sets into pairs (one hypothesis separates only two classes irrespective of the other classes) and then do classification.
- "1-vs-rest" for classification a pair of class and remind classes:
In this strategy, we would divide the data sets such that a hypothesis separates one class label from all of the rest [62].

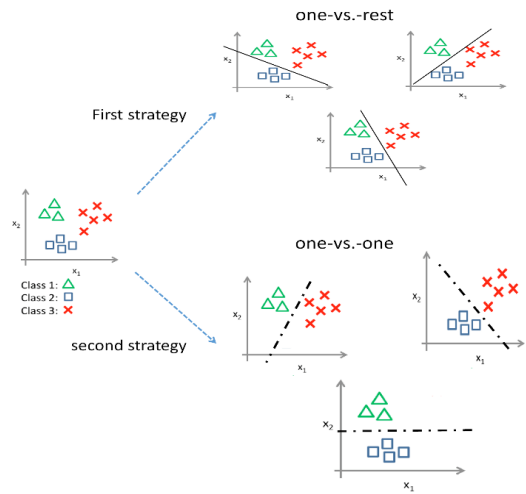

Fig 29: Sample example of a multiclass support vector machine

The SVM algorithms applied to a three-class diagnostic problem are shown in Fig 29.

### C.    Naive Bayes Classifier:

Bayesian theory is one of the statistical methods for classification. Bayesian theory allows a cost function to be introduced for a situation where it is possible that an input vector that is a member of Class A is mistakenly attributed to Class B. In this way, different classes, each in the form of a probability hypothesis, are considered. Each new educational sample increases or decreases the likelihood of previous hypotheses and finally, the assumptions with the highest probability are considered as a class and they will be attacked by them. Bayesian decision theory is a statistical approach to the pattern recognition problem. By using this possible method, we can minimize the error.

$$P(C_i \,|X) \propto \prod_{k=1}^{n} p(x_k|C_i) \times P(C_i) \qquad (39)$$

$$P(x_k|C_i) = N\left(x_k; \mu_{ki}, \sigma_{ki}^2\right) = \frac{1}{\sqrt{2\pi}\,\sigma_{ki}} e^{\frac{-1}{2}\left(\frac{x_{ki}-\mu_{ki}}{\sigma_{ki}}\right)^2} \qquad (40)$$

$$j = arg\,max\,P(C_i|X) \qquad (41)$$

Where, $P$ is the occurrence probability, $\mu_{ki}$ be the mean of the values in $x_k$ associated with class $C_i$, $\sigma_{ki}^2$ is variance of the values in $x_k$ associated with class $C_i$ and $j$ assigns a class label.

The existence articles, where the Naïve Bayes is applied for classification consist of

### D.    Decision Tree:

The Decision tree method creates a tree-based classification model. The decision tree is the tree in which the samples are categorized in some way that they grow from the root to the bottom and eventually reaches the nodes of the leaf. This method is considered as part of the classification method with the supervisor and using the educational samples, can draw a tree called the decision tree based on the characteristics of each of the data or label samples in the test phase and recognize the type of class or label. Via using this tree, rules can be drawn for the inference system and using those label data or samples that do not specify the class or labels. The CLS algorithm was developed by Ross Quilan in 1986 more fully called Inducing Decision trees (ID3). Subsequently, a more complete alternative algorithm, entitled C4.5, addresses some of the deficiencies of the ID3.

*E.     Artificial Neural Network:*

Artificial Neural Networks (ANN) is patterns for information processing constructed on biological neural networks resembling human brain. Pattern identification is the most popular application of neural networks. The basic form of ANN is the multi-layer perceptron (MLP), a neural network that updates weighs through Back-propagation during training.

*F.     Artificial Neural Network:*

Artificial Neural Networks (ANN) is patterns for information processing constructed on biological neural networks resembling human brain. Pattern identification is the most popular application of neural networks. The basic form of ANN is the multi-layer perceptron (MLP), a neural network that updates weighs through Back-propagation during training.

*G.     Other Classifier:*

Other classifiers are also used for texture classification e.g. Decision tress and LSTM (stands for long short-term memory). Decision tress method increase a classification model based on tree. DTs classify samples from rout to down and growth finally to leaves in nods. The LSTM is RNN1 architecture an ANN proposed in 1997[63]. An LSTM network is well-suited to learn from experience to classify when there are very long time lags of unknown size between important events. This is one of the main reasons why LSTM outperforms alternative RNNs and Hidden Markov Models and other sequence learning methods. Short-term memory lasting neural network like a recursive networks, has resolved the problem of forgetting a long sequence data.

# VI.     RESULTS

The properties of texture database and summery of methods with advantages and disadvantages of them as well as accuracy and Dataset of each method are gathered in tables 2 and 3 respectively.

**Table 2: State-of-the-art texture image dataset properties**

| Texture Database | Total sample | Samples in each class | Image size | Number of classes | Criteria | | | |
|---|---|---|---|---|---|---|---|---|
| | | | | | Multi Rotated | Multi scale | Illumination | Viewpoint |
| Brodatz[46] | 112 | 1 | 640×640 | 112 | ✓ | − | − | − |
| VisTex[48] | 640 | 16 | 128×128 | 40 | ✓ | − | ✓ | − |
| Outex(Tc-0010) [47] | 4320 | 180 | 128×128 | 24 | ✓ | − | ✓ | − |
| CUReT[49] | 5612 | 92 | 200×200 | 61 | ✓ | − | ✓ | ✓ |
| UIUC[50] | 1000 | 40 | 640×640 | 25 | ✓ | − | − | ✓ |
| UMD[64] | 1000 | 40 | 960×1280 | 25 | − | − | − | − |
| KTH-TIPS [] | 810 | 81 | 200×200 | 10 | − | ✓ | ✓ | ✓ |
| FMD[53] | 1000 | 100 | − | 10 | − | − | − | − |

---

[1] recurrent neural network

**Table 4: A comparison study on stats-of-the-art texture analysis methods**

| Row | Author | Paper subject | | Dataset | | Type classifier | Accuracy | |
|---|---|---|---|---|---|---|---|---|
| 1 | Ahmadvand and Daliri [24] | Rotation invariant texture classification using extended wavelet channel combining and LL channel filter bank | | VisTex | | KNN | 97.60 | |
| | | | | Outex  - CT10 | | | 99.24 | |
| | | | | Brodatz | | | 98.44 | |
| | | Disadvantages | Higher complexity than LBP in terms of number of extracted features – net presented for color texture images | | | | | |
| | | Advantages | Rotation and illumination invariant | | | | | |
| 2 | Du and Yan[44] | Local spiking pattern and its application | | Outex | Outex 10 | neural network | 86.12 | |
| | | | | | Outex 12 | | 66.62 | |
| | | Disadvantages | Don't extended for color texture images – many input parameters – Don't use color information | | | | | |
| | | Advantages | rotation invariant – Impulse noise Resistant – Illumination invariant | | | | | |
| 3 | Suresh and Shunmuganathan [45] | Gray Level Co-Occurrence Matrix Based Statistical Features | | Brodatz | | KNN | 99.04 | |
| | | Disadvantages | Don't provide a significant reason for selected statistical features – Noise sensitivity – sensitivity   to texture database | | | | | |
| | | Advantages | Better results than PSWT,TSWT and Linear regression – rotation invariant | | | | | |
| 4 | Mehta and Egiazarian[65] | Dominant Rotated Local Binary Patterns | | Outex | Outex 10 | K-NN | 96.26 | |
| | | | | | Outex 12 | | 99.19 | |
| | | | | KTH-TIPS. | | | 96.78 | |
| | | Disadvantages | Impulse noise sensitive – Don't consider color information and global features | | | | | |
| | | Advantages | Complete structural information extracted by LBP- complete structural information extracted by LBP - Lower complexity than some modified LBP versions – Rotation Invariant – Scale resistant | | | | | |
| 5 | Siqueira and et.al[42] | Multi-scale gray level co-occurrence matrices for texture description | | UMD | | KNN | 95.2 | |
| | | | | UIUC | | | 81.7 | |
| | | | | Brodatz | | | 87.2 | |
| | | | | VisTex | | | 96.0 | |
| | | | | Outex | | | 89.4 | |
| | | Disadvantages | noise sensitive – sensitive to the number of histogram bins – Nit adoptable with some classifiers | | | | | |
| | | Advantages | Multi scale – consider color and texture features together | | | | | |
| 6 | Fekri-Ershad[43] | Combination of  Edge& Co-occurrence and Local  Binary Pattern | | Stone | | Naive Bayes | 92 | |
| | | | | | | KNN | 3-NN | 93.3 |
| | | | | | | | 5-NN | 93.6 |
| | | | | | | LADTree | 90 | |

| Row | Author | Paper subject | | Dataset | | Type classifier | Accuracy | |
|---|---|---|---|---|---|---|---|---|
| | | Disadvantages | | High computational complexity | | | | |
| | | Advantages | | Better accuracy than LBP or GLCM | | | | |
| 7 | Guo and et.al[66] | LBP variance (LBPV) with global matching | | Out ex | Outex10 | KNN | | 89.63 |
| | | | | | Outex12 | | | 85.23 |
| | | | | | CUReT | | | |
| | | Disadvantages | | Don't extended for color textures – Noise sensitivity | | | | |
| | | Advantages | | Better Results than many modified LBP versions | | | | |
| 8 | Nguyen and et.al[67] | Statistical binary patterns | | KTH-TIPS | | KNN | | 97.73 |
| | | | | KTH-TIPS 2b | | | | 71.59 |
| | | | | CUReT | | | | 98.73 |
| | | | | UIUC | | | | 97.4 |
| | | | | DTD | | | | 74 |
| | | Disadvantages | | Resolution sensitive – Don't provide global features – High computational complexity | | | | |
| | | Advantages | | rotational invariant – Noise invariant | | | | |
| 9 | Chang and et.al[25] | SVM-PSO based rotation-invariant image texture classification in SVD and DWT domains | | VisTex | | SVM | | 100 |
| | | | | Sun Harvest | | | | 100 |
| | | | | Brodatz | | | | 99.55 |
| | | | | USC | | | | 100 |
| | | | | Coffee beans | | | | 98.22 |
| | | Disadvantages | | Noise sensitivity – High computational complexities than methods in spatial domain | | | | |
| | | Advantages | | Rotation Invariant – Extract features in spatial and frequency domain jointly | | | | |
| 10 | Junior and Backes[68] | ELM based Signature for texture classification | | VisTex | | KNN | | 99.83 |
| | | | | Outex | | | | 97.59 |
| | | | | Brodatz | | | | 99.42 |
| | | Disadvantages | | Many input parameters for neural network ELM – High computational complexity in terms of number of descriptors | | | | |
| | | Advantages | | combination of statistical and structural features | | | | |
| 11 | Hao and et.al[71] | Evaluation of ground distances and features in EMD-based GMM matching | | KTH-TIPS 2b | | KNN | | 78.6 |
| | | | | FMD | | | | 81.7 |
| | | | | ULUC | | | | 84 |
| | | Disadvantages | | High computational complexity – Don't provide reasons for choose distance | | | | |
| | | Advantages | | color and texture analysis jointly – Better Results than GMM based methods | | | | |
| | Fekri-Ershad and Tajeripour[18] | color texture classifications based on proposed impulse noise resistant color local binary patterns and significant point's selection | | Outex TC 13 | | | | 99.83 |
| | | | | KTH-TIPS 2a | | | | 83.57 |
| | | | | VisTex | | | | 98.12 |
| | | Disadvantages | | Higher complexity than original LBP | | | | |
| | | Advantages | | Rotation  invariant – Impulse  noise resistant -   extend LBP method to color textures | | | | |

## VII.    Conclusions:

In this study we try to consider almost all articles which are proposed methods for texture analysis in field of texture classification. As four main categories are defined for texture classification methods, some of primary methods are under unique. However, through expand methods and in novation of combined methods, new extended method of the texture analysis are allocated to more than one category (see Table1). In statistical category  co-occurrence matrix and Local Binary Patten Methods are more popular, and for Model based category the Fractal models is more famous, Gabor and Wavelet are more applicable through Transform based category.

Majority of methods are categorized under statistical and transform based methods or a combination of these method. A main reason of method diversity is change is in way of texture image analyzing (e.g. noise, rotation, scale, illumination, view point). As a result, each new method looking for overcomes some of challenges. Almost all methods are rotation invariant; however, a majority of methods are noise sensitive. For example, DRLBP and LBP variance methods are extend LBP and get better results but those are continuously noise sensitive. Most of methods could applied for gray scale texture images and Fekri-Ershad , Haon and Siqueira have used their own for color texture. Some of methods didn't provide a robust reason for their selections for instance "Gray Level Co-Occurrence Matrix Based Statistical Features".

## REFERENCES

[1] Manjunath, B.S., and Ma, W.Y. (1996). Texture features for browsing and retrieval of image data. *IEEE Transactions on pattern analysis and machine intelligence*, *18*(8), 837-842.

[2] Tuceryan, M., and Jain, A.K. (1993). Texture analysis, in Handbook of pattern recognition and computer vision. *World Scientific Publisher*, 235-276.

[3] Eichkitz, C. G., Davies, J., Amtmann, J., Schreilechner, M.G., and De Groot, P. (2015). Grey level co-occurrence matrix and its application to seismic data, *First Break, 33*(3), 71-77.

[4] Fekri-Ershad, Sh. (2012). Texture classification approach based on energy variation. *International Journal of Multimedia Technology*, *2*(2), 52-55.

[5] Rahim, M.A., Azam, M.S., Hossain, N., and Islam, M.R. (2013). Face recognition using local binary patterns (LBP). *Global Journal of Computer Science and Technology*, 1-12.

[6] Kim, H. I., Lee, S. H., and Ro, Y. M. (2015). Multispectral texture features from visible and near-infrared synthetic face Images for face recognition. Paper presented at the Proceeding of the IEEE International Symposium on Multimedia, pp. 593-596.

[7] Sheltonl, J., Dozier, J., Bryant, K., and Adams, J. (2011). Genetic based LBP feature extraction and selection for facial recognition. Paper presented at the Proceeding of the *49th Annual Southeast Regional Conference*, pp. 197-200.

[8] Zhang, B., Gao, Y., Zhao, S., and Liu, J. (2010). Local derivative pattern versus local binary pattern: face recognition with high-order local pattern descriptor. *IEEE Transactions on Image Processing*, 19(2), 533-544.

[9] Nagy, A.M., Ahmed, A., and Zayed, H. (2014). Particle filter based on joint color texture histogram for object tracking, Paper presented at the Proceeding of the First International Conference on Image Processing, Applications and Systems Conference (IPAS), pp. 1-6.

[10] Raheja, J. L., Kumar, S., and Chaudhary, A. (2013). Fabric defect detection based on GLCM and Gabor filter: A comparison, *Optik-International Journal for Light and Electron Optics*, *124*(23), 6469-6474.

[11] Patterns, M. L. (2017). Breast density classification using multiresolution local quinary patterns in Mammograms, Paper presented at the Proceeding of the 21st Annual Conference in Medical Image Understanding and Analysis, pp. 365-376, Edinburgh, UK.

[12] Rastghalam, R., and Pourghassem, H. (2016). Breast cancer detection using MRF-based probable texture feature and decision-level fusion-based classification using HMM on thermography images, *Pattern Recognition, 51*, 176-186.

[13] Duque, J. C., Patino, J. E., Ruiz, L. A., and Pardo-Pascual, J. E. (2015). Measuring intra-urban poverty using land cover and texture metrics derived from remote sensing data, *Landscape and Urban Planning, 135*, 11-21.

[14] Wood, E. M., Pidgeon, A. M., Radeloff, V. C., & Keuler, N. S. (2012). Image texture as a remotely sensed measure of vegetation structure. *Remote Sensing of Environment*, *121*, 516-526.

[15] Zhang, L., Zhou, Z., & Li, H. (2012, September). Binary gabor pattern: An efficient and robust descriptor for texture classification. In *2012 19Th IEEE international conference on image processing* (pp. 81-84). Ieee.

[16] Tan, X., & Triggs, B. (2007, October). Fusing Gabor and LBP feature sets for kernel-based face recognition. In *International Workshop on Analysis and Modeling of Faces and Gestures* (pp. 235-249). Springer, Berlin, Heidelberg.

[17] Yang, P., & Yang, G. (2016). Feature extraction using dual-tree complex wavelet transform and gray level co-occurrence matrix. *Neurocomputing*, *197*, 212-220.

[18] Fekri-Ershad, Sh., and Tajeripour, F. (2017). Color texture classification based on proposed impulse-noise resistant color local binary patterns and significant points selection algorithm. *Sensor Review, 37*(1), 33-42.

[19] Ershad, S. F. (2011). Color texture classification approach based on combination of primitive pattern units and statistical features. *International Journal of Multimedia & Its Applications*, *3*(3), 1-13.

[20] Liang, H., & Weller, D. S. (2016, September). Edge-based texture granularity detection. In *2016 IEEE International Conference on Image Processing (ICIP)* (pp. 3563-3567). IEEE.

[21] Xu, Y., Yang, X., Ling, H., & Ji, H. (2010, June). A new texture descriptor using multifractal analysis in multi-orientation wavelet pyramid. In *2010 IEEE Computer Society Conference on Computer Vision and Pattern Recognition* (pp. 161-168). IEEE.

[22] Cross, G. R., & Jain, A. K. (1983). Markov random field texture models. *IEEE Transactions on Pattern Analysis & Machine Intelligence*, (1), 25-39.

[23] Mao, J., & Jain, A. K. (1992). Texture classification and segmentation using multiresolution simultaneous autoregressive models. *Pattern recognition*, *25*(2), 173-188.

[24] Ahmadvand, A., & Daliri, M. R. (2016). Rotation invariant texture classification using extended wavelet channel combining and LL channel filter bank. *Knowledge-Based Systems*, *97*, 75-88.

[25] Chang, B. M., Tsai, H. H., & Yen, C. Y. (2016). SVM-PSO based rotation-invariant image texture classification in SVD and DWT domains. *Engineering Applications of Artificial Intelligence*, *52*, 96-107.

[26] Haralick, R. M., & Shanmugam, K. (1973). Textural features for image classification. *IEEE Transactions on systems, man, and cybernetics*, (6), 610-621.

[27] Ojala, T., Pietikäinen, M., & Harwood, D. (1996). A comparative study of texture measures with classification based on featured distributions. *Pattern recognition*, *29*(1), 51-59.

[28] Pietikäinen, M., Ojala, T., & Xu, Z. (2000). Rotation-invariant texture classification using feature distributions. *Pattern Recognition*, *33*(1), 43-52.

[29] Ojala, T., Pietikäinen, M., & Mäenpää, T. (2002). Multiresolution gray-scale and rotation invariant texture classification with local binary patterns. *IEEE Transactions on Pattern Analysis & Machine Intelligence*, (7), 971-987.

[30] Sobel, I. (1990). An isotropic 3 image gradient operator, Machine Vision for Three-Dimensional Scenes (H. Freeman ed.).

[31] Prewitt, J. M. (1970). Object enhancement and extraction. *Picture processing and Psychopictorics*, *10*(1), 15-19.

[32] Lowe, D. G. (1999, September). Object recognition from local scale-invariant features. In *iccv* (Vol. 99, No. 2, pp. 1150-1157).

[33] Lowe, D. G. (2004). Distinctive image features from scale-invariant keypoints. *International journal of computer vision*, *60*(2), 91-110.

[34] Mandelbrot, B. B. (1977). Ebookstore Release Benoit B Mandelbrot Fractals: Form, Chance And Dimension prc, Freeman, p. 365.

[35] Scheunders, P., Livens, S., Van-de-Wouwer, G., Vautrot, P., and Van-Dyck, D. (1998). Wavelet-based texture analysis. *International Journal on Computer Science and Information Management*, *1*(2), 22-34.

[36] Shen, L., & Yin, Q. (2009). Texture classification using curvelet transform. In *Proceedings. The 2009 International Symposium on Information Processing (ISIP 2009)* (p. 319). Academy Publisher.

[37] Arivazhagan, S., Ganesan, L., & Kumar, T. S. (2006). Texture classification using ridgelet transform. *Pattern Recognition Letters*, *27*(16), 1875-1883.

[38] Idrissa, M., & Acheroy, M. (2002). Texture classification using Gabor filters. *Pattern Recognition Letters*, *23*(9), 1095-1102.

[39] Gonzalez, R. C., Woods, R. E., & Eddins, S. L. (1992). Representation and description. *Digital Image Processing*, *2*, 643-692.

[40] Jain, A. K., & Farrokhnia, F. (1991). Unsupervised texture segmentation using Gabor filters. *Pattern recognition*, *24*(12), 1167-1186.

[41] Daugman, J. G. (1988). Complete discrete 2-D Gabor transforms by neural networks for image analysis and compression. *IEEE Transactions on acoustics, speech, and signal processing*, *36*(7), 1169-1179.

[42] De Siqueira, F. R., Schwartz, W. R., & Pedrini, H. (2013). Multi-scale gray level co-occurrence matrices for texture description. *Neurocomputing*, *120*, 336-345.

[43] Ershad, S. F. (2011). Texture Classification Approach Based on Combination of Edge & Co-occurrence and Local Binary Pattern. In *Proceedings of the International Conference on Image Processing, Computer Vision, and Pattern Recognition (IPCV)* (p. 1). The Steering Committee of The World Congress in Computer Science, Computer Engineering and Applied Computing (WorldComp).

[44] Du, S., Yan, Y., & Ma, Y. (2016). Local spiking pattern and its application to rotation-and illumination-invariant texture classification. *Optik*, *127*(16), 6583-6589.

[45] Suresh, A., & Shunmuganathan, K. L. (2012). Image texture classification using gray level co-occurrence matrix based statistical features. *European Journal of Scientific Research*, *75*(4), 591-597.

[46] P. Brodatz. Textures: A Photographic Album for Atists and Designers [Online]. Available: http://multibandtexture.recherche.usherbrooke.ca/colored%20_brodatz.html

[47] Outex Texture Database, 2007-10-01 ed. http://www.outex.oulu.fi/index.php?page=classification: University of Oulu.

[48] Vision Texture (VisTex Database), ed. http://vismod.media.mit.edu/vismod/imagery/VisionTexture/vistex.html: Massachusetts Institute of Technology., 2002.

[49] Columbia-Utrecht Reflectance and Texture Database, ed. http://www1.cs.columbia.edu/CAVE//exclude/curet/.index.html: Columbia University and Utrecht University.

[50] UIUC Image Database, ed. http://www-cvr.ai.uiuc.edu/ponce_grp/data/.

[51] KTH-TIPS,ed. http://www.nada.kth.se/cvap/databases/kth-tips/.

[52] The KTH-TIPS and KTH-TIPS2 image databases, 2006-06-09 ed. http://www.nada.kth.se/cvap/databases/kth-tips/index.html.

[53] S. Lavanya, R. Rosenholtz, and E. Adelson, Flickr Material Database (FMD), ed. https://people.csail.mit.edu/celiu/CVPR2010/FMD/, 2014.

[54] G. Kylberg, Kylberg Texture Dataset v. 1.0, 2014-03-26 ed. www.cb.uu.se/~gustaf/texture/: Uppsala University, 2011.

[55] S. F. a. M. J. H. Renaud Péteri. The DynTex database Homepage [Online]. Available: http://dyntex.univ-lr.fr/classification_datasets/classification_datasets.html

[56] TILDA Textile Texture Database, 2011 – 2017 ed. https://lmb.informatik.uni-freiburg.de/resources/datasets/tilda.en.html: LMB, University of Freiburg.

[57] Describable Textures Dataset (DTD), 2012 ed. https://www.robots.ox.ac.uk/~vgg/data/dtd/: Johns Hopkins Centre for Language and Speech Processing (CLSP).

[58] The Photex Database, 2011 ed. http://www.macs.hw.ac.uk/texturelab/resources/databases/photex/: The Texturelab Edinburgh.

[59] D. Chugai. Texture library [Online]. Available: http://texturelib.com/

[60] Cortes, C., & Vapnik, V. (1995). Support-vector networks. *Machine learning*, *20*(3), 273-297.

[61] Vapnik, V., & Mukherjee, S. (2000). Support vector method for multivariate density estimation. In *Advances in neural information processing systems* (pp. 659-665).

[62] Platt, J. C., Cristianini, N., & Shawe-Taylor, J. (2000). Large margin DAGs for multiclass classification. In *Advances in neural information processing systems* (pp. 547-553).

[63] Hochreiter, S., & Schmidhuber, J. (1997). Long short-term memory. *Neural computation*, *9*(8), 1735-1780.

[64] Viewpoint invariant texture description (UMD Texture Dataset), ed. http://legacydirs.umiacs.umd.edu/~fer/website-texture/texture.htm.

[65] Mehta, R., & Egiazarian, K. (2016). Dominant rotated local binary patterns (DRLBP) for texture classification. *Pattern Recognition Letters*, *71*, 16-22.

[66] Guo, Z., Zhang, L., and Zhang, D. (2010). Rotation invariant texture classification using LBP variance (LBPV) with global matching, *Pattern recognition, 43*(3), 706-719.

[67] Nguyen, T.P., Vu, N. S., and Manzanera, A., (2016). Statistical binary patterns for rotational invariant texture classification, *Neurocomputing, 173*, 1565-1577.

[68] Junior, J.J., and Backes, A. R. (2016). ELM based signature for texture classification. *Pattern Recognition, 51*, 395-401.

[69] Tajeripour, F., Saberi, M., Rezaei, M., and Fekri-Ershad, Sh. (2011) Texture classification approach based on combination of random threshold vector technique and co-occurrence matrixes. Paper presented at the Proceeding of the International Conference on Computer Science and Network Technology (ICCSNT), Vol. 4, pp. 2303-2306, Harbin-China.

[70] Fekri-Ershad, Sh., and Tajeripour, F., (2017). Impulse-Noise resistant color-texture classification approach using hybrid color local binary patterns and kullback–leibler divergence. *Computer Journal*, *60*(11), 1633-1648.

[71] Hao, H., Wang, Q., Li, P., and Zhang, L. (2016). Evaluation of ground distances and features in EMD-based GMM matching for texture classification, *Pattern Recognition*, *57*, 152-163.

[72] Tajeripour, F., & Fekri-Ershad, S. H. (2012, February). Porosity detection by using improved local binary pattern. In *Proceedings of the 11th WSEAS International Conference on Signal Processing, Robotics and Automation (ISPRA'12)* (Vol. 1, pp. 116-121).

[73] Armi, L., & Fekri-Ershad, S. (2019). Texture image Classification based on improved local Quinary patterns. *Multimedia Tools and Applications*, 1-24.

## Authors Biography

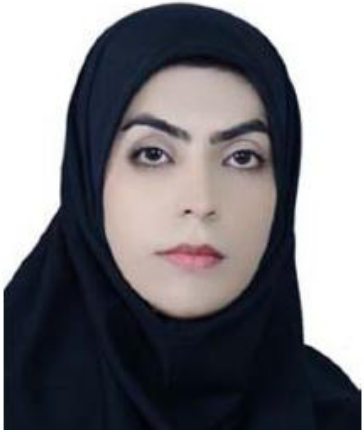

**Laleh Armi** received her M.Sc. majored in computer engineering-artificial intelligence from najafabad branch, Islamic azad university in 2017. Her research interests are image processing and computer vision applications includes texture analysis, texture classification, image descriptors, etc.

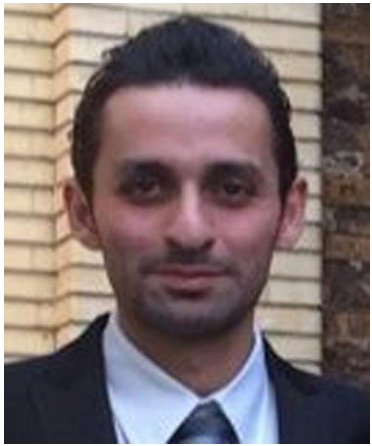

**Shervan Fekri-Ershad** received his B.Sc. degree in computer hardware engineering from Islamic Azad University of Najaf-Abad, Iran in 2009. He received his M.Sc. & Ph.D. degrees from International Shiraz University, Iran in 2012–2016, majored in Artificial Intelligence. He joined the faculty of computer engineering at Najafabad branch, Islamic azad university, Najafabad, Iran as a staff member (assistant professor) in 2012. His research interests are image processing applications includes visual inspection systems, visual classification, texture analysis, surface defect detection and etc.